\documentclass[10pt,twocolumn,letterpaper]{article}

\usepackage[pagenumbers]{cvpr} 

\usepackage{pifont}
\usepackage{colortbl}
\usepackage{rotating} 
\usepackage{multirow}
\usepackage{marvosym}

\usepackage[dvipsnames]{xcolor}

\definecolor{cvprblue}{rgb}{0.21,0.49,0.74}
\usepackage[pagebackref,breaklinks,colorlinks,citecolor=cvprblue]{hyperref}

\title{Point Cloud Pre-training with Diffusion Models}

\newcommand{\ourmethod}{PointDif\xspace}

\author{
    Xiao Zheng\textsuperscript{\rm 1} \quad
    Xiaoshui Huang\textsuperscript{\rm 2}\thanks{Corresponding authors}\quad
    Guofeng Mei\textsuperscript{\rm 3} \quad
    Yuenan Hou\textsuperscript{\rm 2} \\
    Zhaoyang Lyu\textsuperscript{\rm 2}  \quad
    Bo Dai\textsuperscript{\rm 2} \quad
    Wanli Ouyang\textsuperscript{\rm 2} \quad
    Yongshun Gong\textsuperscript{\rm 1}\footnotemark[1] \\
    \textsuperscript{\rm 1}Shandong University \quad
    \textsuperscript{\rm 2}Shanghai AI Laboratory \quad
    \textsuperscript{\rm 3}Fondazione Bruno Kessler
    }

\begin{document}
\maketitle

\begin{abstract}
Pre-training a model and then fine-tuning it on downstream tasks has demonstrated significant success in the 2D image and NLP domains. However, due to the unordered and non-uniform density characteristics of point clouds, it is non-trivial to explore the prior knowledge of point clouds and pre-train a point cloud backbone. In this paper,  we propose a novel pre-training method called \textbf{Point} cloud \textbf{Dif}fusion pre-training (\ourmethod). We consider the point cloud pre-training task as a conditional point-to-point generation problem and introduce a conditional point generator. This generator aggregates the features extracted by the backbone and employs them as the condition to guide the point-to-point recovery from the noisy point cloud, thereby assisting the backbone in capturing both local and global geometric priors as well as the global point density distribution of the object. We also present a recurrent uniform sampling optimization strategy, which enables the model to uniformly recover from various noise levels and learn from balanced supervision. Our \ourmethod achieves substantial improvement across various real-world datasets for diverse downstream tasks such as classification, segmentation and detection. Specifically, \ourmethod~attains \textbf{70.0\%} mIoU on S3DIS Area 5 for the segmentation task and achieves an average improvement of \textbf{2.4\%} on ScanObjectNN for the classification task compared to TAP. Furthermore, our pre-training framework can be flexibly applied to diverse point cloud backbones and bring considerable gains.
\end{abstract}
\section{Introduction}

\begin{figure}[t]
    \centering
    \includegraphics[width=0.49\textwidth]{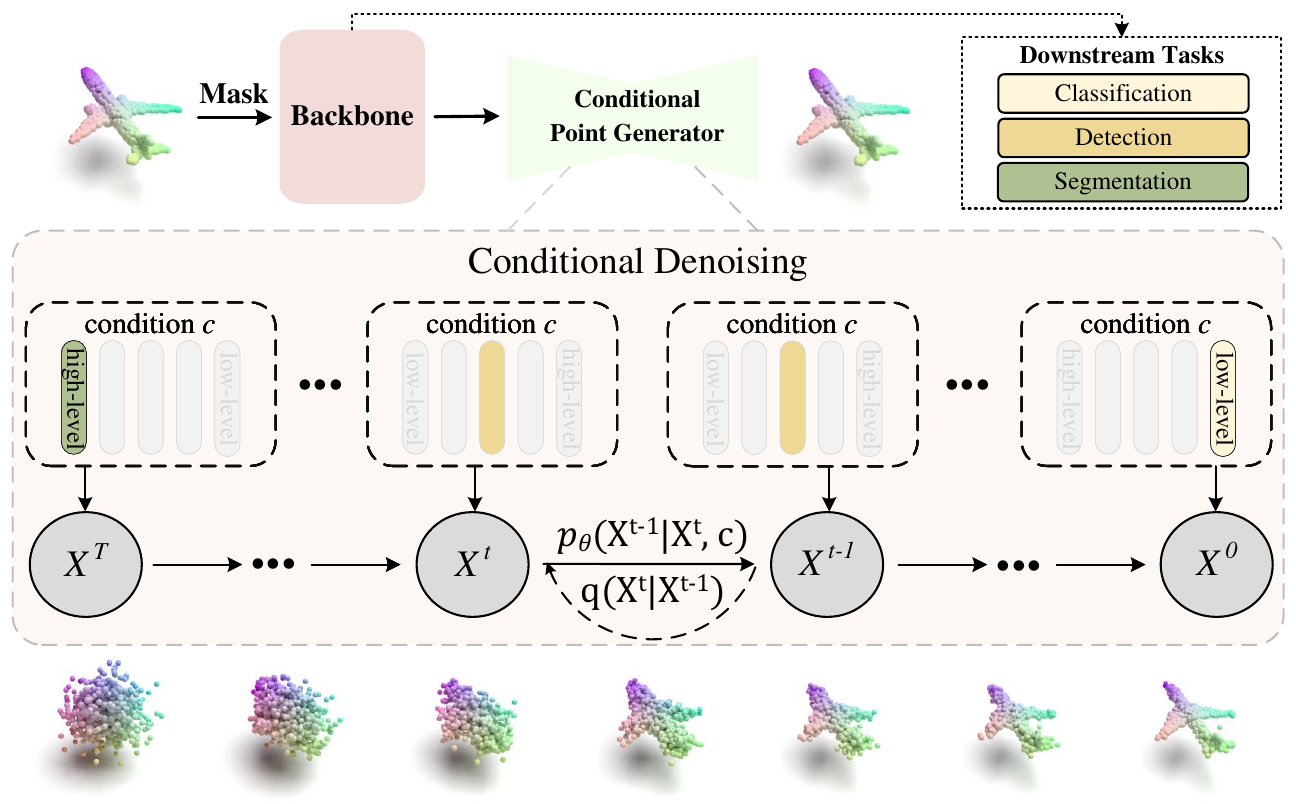}
    \captionsetup{skip=1pt}
    \caption{\textbf{Schematic illustration of our \ourmethod.}  Our \ourmethod can pre-train different backbones by reconstructing the original point cloud point-to-point from the noisy point cloud. During pre-training, the latent features guide the restoration of noisy point clouds at various levels, allowing the backbone to learn more hierarchical geometric prior.} 
    \label{fig:schematic}
    \vspace{-0.5cm}
\end{figure}

In recent years, a surging number of studies, including SAM~\cite{kirillov2023segment}, VisualChatGPT~\cite{wu2023visual}, and BLIP-2 \cite{li2023blip}, have demonstrated the exceptional performance of pre-trained models across a broad range of 2D image and natural language processing (NLP) tasks. Pre-training on large-scale datasets endows the model with abundant prior knowledge, enabling the pre-trained models to exhibit superior performance and enhanced generalization capabilities after finetuning, compared to models trained solely on downstream tasks \cite{huang2022frozen,pang2022masked,li2023blip}. Similar to the 2D and NLP fields, pre-training methods in point cloud data have also become essential in enhancing model performance and boosting model generalization ability.

Contemporary point cloud pre-training methods can be casted into two categories, \ie, contrastive-based and generative-based pre-training. Contrastive-based methods~\cite{xie2020pointcontrast,zhang2021self,afham2022crosspoint} resort to the contrastive objective to make deep models grasp the similarity knowledge between samples. By contrast, generative-based methods involve pre-training by reconstructing the masked point cloud~\cite{pang2022masked, zhang2022point} or its 2D projections~\cite{wang2023take, huang2023ponder}. However, several factors mainly account for the inferior pre-training efficacy in the 3D domain. For contrastive-based methods \cite{xie2020pointcontrast,afham2022crosspoint}, selecting the proper negative samples to construct the contrastive objective is non-trivial. The generative-based pre-training approaches, such as Point-MAE~\cite{pang2022masked} and Point-M2AE~\cite{zhang2022point}, solely reconstruct the masked point patches.  In this way, they cannot capture the global density distribution of the object. Additionally, there is no precise one-to-one matching for MSE loss and set-to-set matching for Chamfer Distance loss between reconstructed and original point cloud due to its unordered nature. Besides, the projection from 3D to 2D by TAP~\cite{wang2023take} and Ponder \cite{huang2023ponder} inevitably introduces the geometric information loss, making the reconstruction objective difficult to equip the backbone with comprehensive geometric prior.

To combat against the unordered and non-uniform density characteristics of point clouds, inspired by adding noise and denoising of the diffusion model~\cite{ho2020denoising}, we propose a novel diffusion-based pre-training framework, dubbed \ourmethod. It pre-trains the point cloud backbone by restoring the noisy data at each step as illustrated in \cref{fig:schematic}. This procedural denoising process is similar to the visual streams in our human brain mechanism~\cite{takagi2022high}. Human uses this simple brain mechanism to obtain broad prior knowledge from the 3D world. Similarly, we find that low-level and high-level neural representation emerges from denoising neural networks. This aligns with our goal of applying pre-trained models to downstream low-level and high-level tasks, such as classification and segmentation. Moreover, the diffusion model has strong theoretical guarantees and provides an inherently hierarchical learning strategy by enabling the understanding of data distribution hierarchically.

Specifically, we present a conditional point generator in our \ourmethod, which guides the point-to-point generation from the noisy point cloud. This conditional point generator encompasses a Condition Aggregation Network (CANet) and a Conditional Point Diffusion Model (CPDM). The CANet is responsible for globally aggregating latent features extracted by the backbone. The aggregated features serve as the condition to guide the CPDM in denoising the noisy point cloud. During the denoising process, the point-to-point mapping relationship exists in the noisy point cloud at neighboring time steps. Equipped with the CPDM, the backbone can effectively capture the global point density distribution of the object. This enables the model to adapt to downstream tasks that involve point clouds with diverse density distributions. With the help of the conditional point generator, our pre-training framework can be extended to various point cloud backbones and enhance their overall performance.

Moreover, as shown in \cref{tab:time_intervals}, we find that sampling time step $t$ from different intervals during pre-training can learn different levels of geometric prior. Based on this observation, we propose a recurrent uniform sampling optimization strategy. This strategy divides the diffusion time steps into multiple intervals and uniformly samples the time step \(t\) throughout the pre-training process. In this way, the model can uniformly recover from various noise levels and learn from balanced supervision.  To the best of our knowledge, we are the first to demonstrate the effectiveness of generative diffusion models in enhancing point cloud pre-training.

Our key contributions can be summarized as follows:
\begin{itemize}[noitemsep,topsep=0pt,leftmargin=*]

\item We propose the first framework for point cloud pre-training based on diffusion models, called \ourmethod. Performing iterative denoising on the noisy point cloud can assist backbones in acquiring a comprehensive understanding of the original point cloud and extracting hierarchical geometric prior.

\item We present a conditional point generator to guide the point-to-point generation from the noisy point cloud. This facilitates the network in capturing the global point density distribution of the object.

\item We introduce a recurrent uniform sampling strategy that assists the model in uniformly restoring diverse noise levels and learning from balanced supervision. 

\item Our \ourmethod demonstrates competitive performance across various real-world downstream tasks. Furthermore, our framework can be flexibly applied to diverse point cloud backbones and enhance their performance.

\end{itemize}
\section{Related Work}
\label{sec:related}

\begin{figure*}[t]
    \centering
    \includegraphics[width=0.90\textwidth]{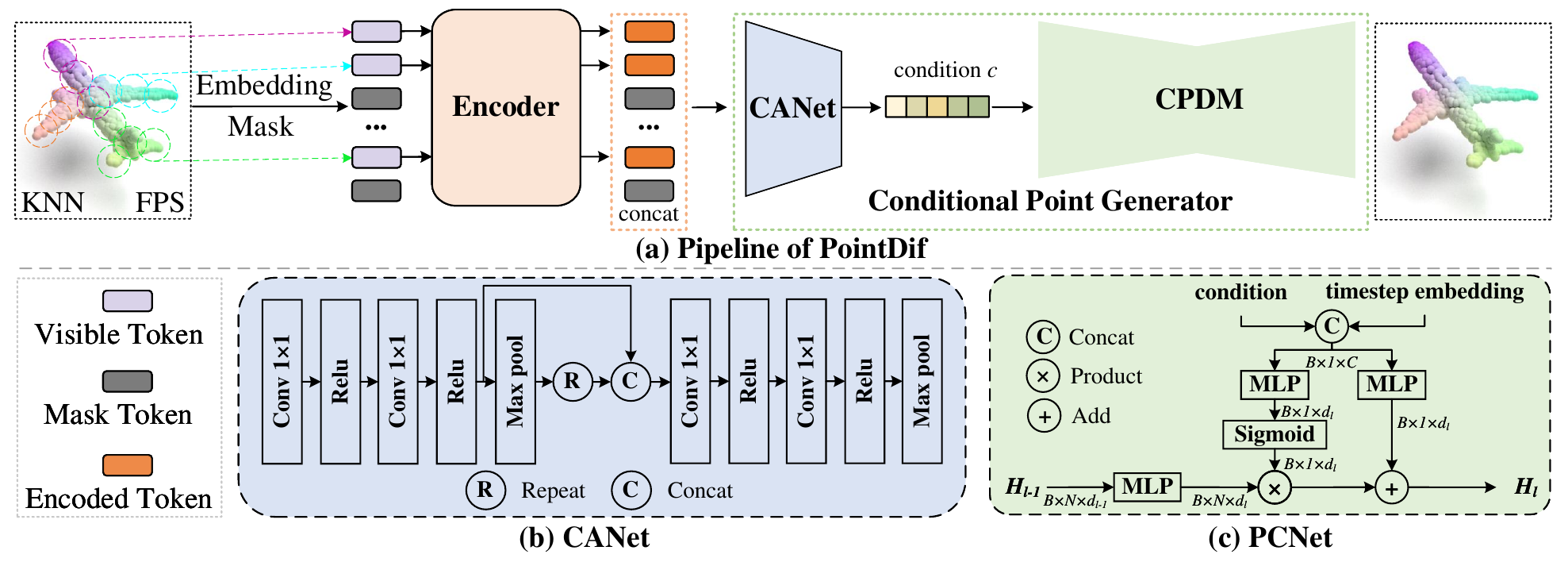}
    \captionsetup{skip=1pt}
    \caption{ (a) The pipeline of our \ourmethod. We first divide the input point cloud into masked and embedded point patches. Then, a transformer encoder is used to extract the latent features. Finally, we employ the condition aggregation network (CANet) to aggregate latent features to obtain the condition \(c\), and then guide the conditional point diffusion model (CPDM) to point-to-point recovery of the original point cloud from the randomly perturbed point cloud. (b) The detailed structure of CANet. (c) The detailed structure of the point condition network (PCNet), CPDM is composed of six PCNet.}
    \label{fig:pipeline}
    \vspace{-0.5cm}
\end{figure*}

This section first briefly reviews existing point cloud pre-training approaches. Since the diffusion model is a primary component in the proposed pre-training framework, we also review the relevant studies on diffusion models.

\noindent \textbf{Pre-training for 3D point cloud.}
Contrastive-based algorithms pre-train the backbone by comparing the similarities and differences among samples. PointContrast~\cite{xie2020pointcontrast} is the pioneering method, which constructs two point clouds from different perspectives and compares point feature similarities for point cloud pre-training. Recent research efforts have improved network performance through data augmentation~\cite{zhang2021self, wu2023masked} and the introduction of cross-modal information~\cite{afham2022crosspoint, huang2023clip2point, zeng2023clip2}. In contrast, generative-based pre-training methods focus on pre-training the encoder by recovering masked information or its 2D projections. Point-BERT~\cite{Point-BERT} and Point-MAE~\cite{pang2022masked} respectively incorporate the ideas of BERT~\cite{devlin2018bert}  and MAE~\cite{he2022masked} into point cloud pre-training. TAP~\cite{wang2023take} and Ponder \cite{huang2023ponder} pre-train the point cloud backbone by generating the 2D projections of the point cloud. Point-M2AE~\cite{zhang2022point} constructs a hierarchical network capable of gradually modeling geometric and feature information. Joint-MAE~\cite{guo2023joint} focuses on the correlation between 2D images and 3D point cloud and introduces hierarchical modules for cross-modal interaction to reconstruct masked information for both modalities. Compared to the architectural improvements made in Point-M2AE and Joint-MAE, our method concentrates on refining the training approach. Our \ourmethod leverages the progressive guidance characteristic of the conditional diffusion model, allowing the backbone to learn hierarchical geometric prior by restoring noisy point clouds at different noise levels.

\noindent \textbf{Diffusion Probabilistic Models.}
The diffusion model is inspired by the principles of non-equilibrium thermodynamics and leverages the diffusion process and noise reduction to generate high-quality data. It has shown excellent performance in both generation effectiveness and interpretability. The diffusion model has achieved remarkable success across various domains, including image generation~\cite{nichol2021glide, dhariwal2021diffusion, ramesh2022hierarchical, saharia2022photorealistic, rombach2022high, zhao2023uni} and 3D generation~\cite{lin2023magic3d, xu2023dream3d, Tang_2023_ICCV, wang2023prolificdreamer, liu2023one2345}. Recently, researchers have investigated methods for accelerating the sampling process of DDPM to improve its generation efficiency~\cite{song2020denoising, lu2022dpm, lu2022dpm2}. Moreover, some studies have explored the application of diffusion models in discriminative tasks, such as object detection\cite{chen2022diffusiondet} and semantic segmentation~\cite{amit2021segdiff, wolleb2022diffusion, baranchuk2021label}. 

To our knowledge, we are the first to apply the diffusion model for point cloud pre-training and have achieved promising results. The most relevant work is the 2D pre-training method DiffMAE~\cite{wei2023diffusion}. However, there are four critical distinctions between our \ourmethod~and DiffMAE. Firstly, as to the reconstruction target, DiffMAE pre-trains the network by denoising pixel values of masked patches. In contrast, our \ourmethod pre-trains the network by recovering the original point clouds from randomly noisy point clouds, which is beneficial for the network to learn both local and global geometrical priors of 3D objects. Secondly, as for the guidance way, DiffMAE uses the conditional guidance method of cross-attention. We adopt a point condition network (PCNet) for point cloud data to facilitate 3D generation through point-by-point guidance. It also assists the network in learning the global point density distribution of the object. Thirdly, regarding the loss function, DiffMAE introduces an additional CLIP loss to constrain the model, whereas our \ourmethod demonstrates strong performance in various 3D downstream tasks without additional constraints. Finally, with regard to the unity of the framework, DiffMAE can only pre-train the 2D transformer encoder. In comparison, with the help of our conditional point generator, we can pre-train various point cloud backbones and enhance their performance.
\section{Methodology}

We take pre-training the transformer encoder as an example to introduce our overall pre-training framework, \ie, \ourmethod. The framework can also be easily applied to pre-train other backbones. The pipeline of our \ourmethod is shown in \hyperref[fig:pipeline]{Fig. 2a}.
Given a point cloud, we first divide it into point patches and apply embedding and random masking operations to each patch. Subsequently, we use a transformer encoder to process visible tokens to learn the latent features, which are then used to generate the condition \(c\) through the CANet. Finally, this condition gradually guides the CPDM to recover the original input point cloud from the random noise point cloud in a point-to-point manner. We \emph{pre-train the transformer encoder} to acquire the hierarchical geometric prior through the progressively guided process.

\subsection{Preliminary: Conditional Point Diffusion}

During the diffusion process, random noise is continuously introduced into the point cloud through a Markov chain, and there exists a point-to-point mapping relationship between noisy point clouds of adjacent timestamps. Formally, given a clean point cloud \( X^{0} \in \mathbb{R} ^ {n \times 3} \) containing \( n \) points from the real data distribution \(p_{data} \), the diffusion process gradually adds Gaussian noise to \(X^{0}\) for \(T\) time steps:
\vspace{-0.3cm}
 \begin{equation}
 \small
q(X^{1:T}|X^{0}) = \prod \limits_{t=1}^T q(X^{t}|X^{t-1}),
\end{equation}

\vspace{-0.5cm}
 \begin{equation}
 \small
 \text{where }q(X^{t}|X^{t-1}) = \mathcal{N}(X^{t};\sqrt{1-\beta_{t}}X^{t-1},\beta_{t}I),
\end{equation}

\noindent the hyperparameters \(\beta_{t}\) are some pre-defined small constants and gradually increase over time. \(X^{t}\) is sampled from a Gaussian distribution with mean \(\sqrt{1-\beta_{t}}X^{t-1}\) and variance \(\beta_{t}I\). Moreover, according to ~\cite{ho2020denoising}, it is possible to elegantly express \(X^{T}\) as a direct function of \(X^{0}\):

\vspace{-0.3cm}
\begin{equation}
\small
q(X^{t}|X^{0}) = \mathcal N(X^{t};\sqrt{\bar{\alpha}_{t}}X^{0},(1-\bar{\alpha}_{t})I),
\end{equation}
\vspace{-0.3cm}

\noindent where \(\bar{\alpha}_{t} = \prod_{i=1}^t \alpha_{i}\) and \(\alpha_{t} = 1-\beta_{t} \). As the time step \(t\) increases, \(\bar{\alpha}_{t}\) gradually approaches 0 and \(q(X^{t}|X^{0})\) will be close to the Gaussian distribution \(p_{noise}\).

The reverse process involves using a neural network parameterized by \(\theta\) to gradually denoise a Gaussian noise into a clean point cloud with the help of the condition \(c\). This process can be defined as:

\vspace{-0.3cm}
\begin{equation}
\small
p_{\theta}(X^{0:T},c) = p(X^{T})\prod \limits_{t=1}^T p_{\theta}(X^{t-1}|X^{t},c),
\end{equation}
\vspace{-0.3cm}

\vspace{-0.6cm}
\begin{equation}
\small
\text{where } p_{\theta}(X^{t-1}|X^{t},c) = \mathcal N(X^{t-1}; \mu_{\theta}(X^{t},t,c), \sigma_{t}^{2}I),
\end{equation}
\vspace{-0.5cm}

\noindent the \(\mu_{\theta}\) is a neural network that predicts the mean, and \(\sigma_{t}^{2}\) is a constant that varies with time.

The training objective of the diffusion model is formulated based on variational inference, which employs the variational lower bound (\(vlb\)) to optimize the negative log-likelihood:

\setlength{\belowdisplayskip}{-100pt}
\vspace{-0.6cm}
\begin{equation} \label{eq6}
\small
\begin{split}
L_{vlb} &= E_{q}[-\text{log} p_{\theta}(X^{0}|X^{1},c)+ D_{\text{KL}}(q(X^{T}|X^{0})||p(X^{T})) \\ &+\sum\limits_{t=2}\limits^{T}D_{\text{KL}}(q(X^{t-1}|X^{t},X^{0})||p_{\theta}(X^{t-1}|X^{t},c))],
\end{split}
\end{equation}
\vspace{-0.4cm}

\noindent where \(D_{\text{KL}}(\cdot)\) is the KL divergence. However, training \(L_{vlb}\) is prone to instability. To address this, we adopt a simplified version of the mean squared error~\cite{ho2020denoising}:

\vspace{-0.6cm}
\begin{equation}\label{eq7}
\small
L({\theta})=\mathbb{E}_{t,X^{0},c,\epsilon} \left[\|{\epsilon}-\epsilon_{{\theta}}({\sqrt{{\bar{\alpha}}_{t}}}{{X}}^{0}+{\sqrt{1-{\bar{\alpha}}_{t}}}{{\epsilon}},{{c}},t)\|^{2}\right],
\end{equation}
\vspace{-0.6cm}

\noindent where \({\epsilon} \sim{\mathcal{N}}(0, I) \), \(\epsilon_{{\theta}}(\cdot)\) is a trainable neural network that takes the noisy point cloud \(X^{t}\) at time \(t\), along with the time \(t\) and condition \(c\) as inputs. This network predicts the added noise \(\epsilon\). Additional details regarding derivations and proofs can be found in \cref{sec:proof}.

\subsection{Point Cloud Processing}
The goal of point cloud processing is to convert the given point cloud into several tokens, which consist of point patch embedding and patch masking.

\noindent\textbf{Point Patch Embedding.} 
Following Point-BERT~\cite{Point-BERT} and Point-MAE~\cite{pang2022masked}, we divide the point cloud into point patches using a grouping strategy. Specifically, for an input point cloud \( X \in \mathbb{R} ^ {n \times 3} \) consisting of \( n \) points, we first employ the Farthest Point Sampling (FPS) algorithm to sample \( s \) center points $\{C_{i}\}_{i=1}^s$. For each center point \( C_{i} \), we use the K Nearest Neighborhood (KNN) algorithm to gather the \( k \) nearest points as a point patch \(P_{i}\).

\vspace{-0.3cm}
\begin{equation}
\small
  \{C_{i}\}_{i=1}^s = \text{FPS}(X), \quad \{P_{i}\}_{i=1}^s = \text{KNN}(X, \{C_{i}\}_{i=1}^s).
\end{equation}

It is noteworthy that we apply a centering process to the point patches, which involves subtracting the coordinates of the point center from each point within the patch. This operation helps improve the convergence of the model. Subsequently, we utilize a simplified PointNet~\cite{qi2017pointnet} \(\xi_{\phi}(\cdot)\) with parameter \(\phi\), which employs \(1\times1\) convolutions and max pooling, to embed the point patches \(\{P_{i}\}_{i=1}^s\) into tokens \(\{F_{i}\}_{i=1}^s\).

\vspace{-0.4cm}
\begin{equation}
\small
\{F_{i}\}_{i=1}^s = \xi_{\phi}(\{P_{i}\}_{i=1}^s).
\end{equation}
\vspace{-0.4cm}

\noindent\textbf{Patch Masking.} In order to preserve the geometric information within the patch, we randomly mask the entire points in the patch to obtain the masked tokens \(\{F_{i}^m\}_{i=1}^{r}\) and visible tokens \(\{F_{i}^v\}_{i=1}^g\), where \(r {=} \lfloor s {\times} {m} \rfloor \) is the number of masked tokens, 
\(g {=} s{-}r\) is the number of visible tokens, $\lfloor . \rfloor$ is the floor operation and \(m\) denotes the masking ratio. We conduct experiments to assess the impact of different masking ratios and find that higher masking ratios (0.7-0.9) result in better performance, as discussed in \cref{sec:ablation}.

\subsection{Encoder}
The transformer encoder is responsible for extracting latent geometric features, which is retained for feature extraction during fine-tuning for downstream tasks. \(\Phi_{\rho}(\cdot)\) is our encoder with parameter \(\rho\), composed of 12 standard transformer blocks. To better capture meaningful 3D geometric prior, we remove the masked tokens and encode only the visible tokens \(\{F_{i}^v\}_{i=1}^g\). Furthermore, 
we introduce a position embedding \(\psi_\tau(\cdot)\) with parameter \(\tau\) to embed the position information of the point patch into \(Pos^v_{i}\), which is comprised of two learnable MLPs and the GELU activation function. Then, the position embedding output is concatenated with \(F^v_{i}\) and sent through a sequence of transformer blocks for feature extraction.

\vspace{-0.3cm}
\begin{equation}
\small
\{{T^{v}_{i}}\}_{i=1}^g = \Phi_{\rho}(\{\text{Concat}(F_{i}^v,Pos_{i}^v)\}_{i=1}^g),
\end{equation}
\vspace{-0.3cm}

\vspace{-0.3cm}
\begin{equation}
\small
\text{where } \{Pos^{v}_{i}\}_{i=1}^g = \psi_\tau(\{C^{v}_{i}\}_{i=1}^g).
\end{equation}
\vspace{-0.3cm}

\subsection{Conditional Point Generator}

Our conditional point generator consists of the CANet and the CPDM.

\noindent\textbf{Condition Aggregation Network (CANet).}
To be specific, we concatenate features \(\{T^v_{i}\}_{i=1}^g\) of the visible patches extracted by the encoder with a set of learnable masked patch information \(\{T^m_{i}\}_{i=1}^r\), while preserving their original position information. Afterward, the concatenated features are encoded using the CANet, denoted as \(f_{\omega}(\cdot)\) with the parameter \(\omega\). As shown in \hyperref[fig:pipeline]{Fig. 2b}, our CANet consists of four 1x1 convolutional layers and two global max-pooling layers to aggregate the global contextual features of the point cloud. Ultimately, this process yields the guiding condition \(c\) required for the CPDM:

\begin{figure}[t]
    \centering
    \includegraphics[width=0.48\textwidth]{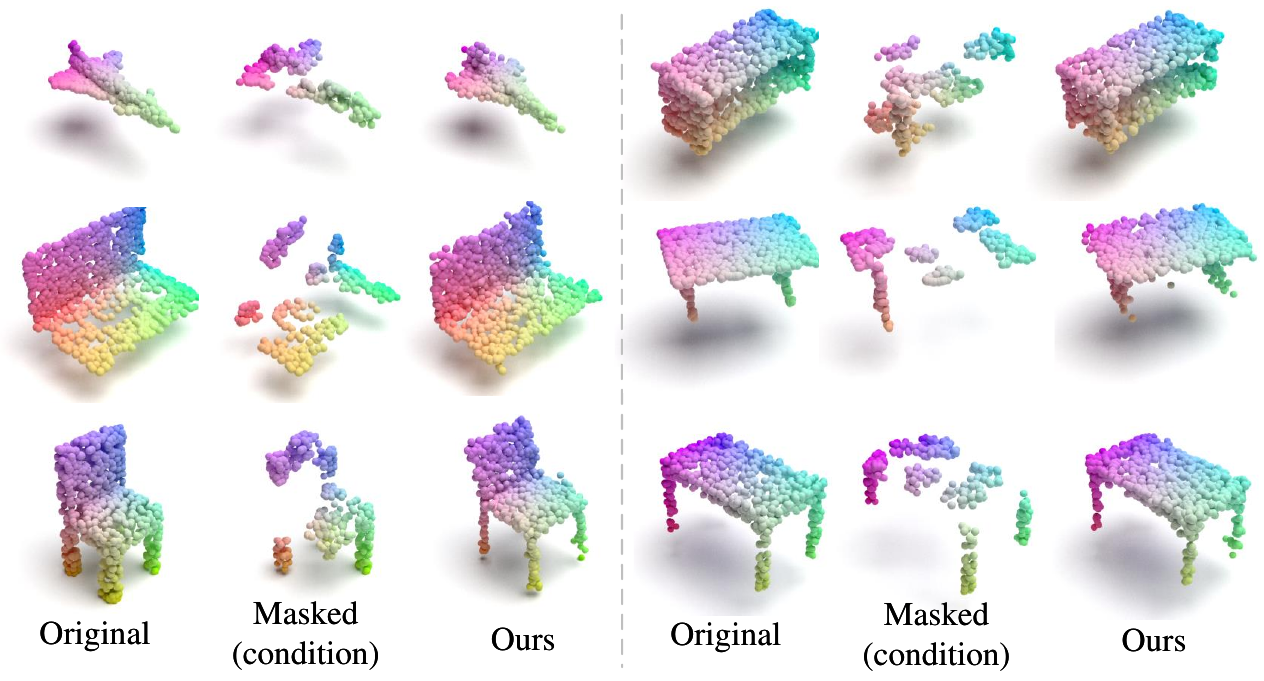}
    \captionsetup{skip=2.0pt}
    \caption{\textbf{Visualization results on the ShapeNet validation set.} Each row visualizes the input point cloud, masked point cloud, and reconstructed point cloud. Even though we mask 80\% points, \ourmethod still produces high-quality point clouds.}
    \label{fig:visualization}
    \vspace{-0.5cm}
\end{figure}

\vspace{-0.3cm}
\begin{equation}
\small
c = f_{\omega}(\text{Concat}(\{T^v_{i}\}_{i=1}^g,\{T^m_{i}\}_{i=1}^r)\}).
\end{equation}
\noindent\textbf{Conditional Point Diffusion Model (CPDM).}
Inspired by~\cite{luo2021diffusion}, we adopt a point diffusion model, which utilizes the condition to guide the recovery of the original point cloud from a randomly perturbed point cloud in a point-to-point way. As illustrated in \hyperref[fig:pipeline]{Fig. 2c}, the conditional point diffusion model comprises six point condition network (PCNet).  The specific structure of each PCNet can be represented as follows:

\vspace{-0.5cm}
\begin{equation}
\small
H_{l}=R_{l}\odot(W_{lh}H_{l-1}+b_{lh})+W_{lb}y, \ \  R_{l}=\sigma(W_{lr}y+b_{lr}),
\end{equation}
where \(H_{l-1}\) and \(H_{l}\) are respectively the input and output of PCNet, \(\sigma\) represents the sigmoid function, and \(W_{l*}, b_{l*}\) are all trainable parameters. \(y\) represents the feature obtained by concatenating the condition \(c\) with the time step embedding. The input dimensions for each PCNet are [3, 128, 256, 512, 256, 128] and the output dimension of the last PCNet is 3. By incorporating the condition into the control mechanism of the reset gate \(R_{l}\), the model can adaptively select geometric features to denoise. 
Recovering from noisy point clouds through point-to-point guidance can aid the network in learning the overall point density distribution of the object. This, in turn, assists different backbones in learning a broader range of dense and sparse geometric priors, resulting in enhanced performance in downstream tasks related to indoor and outdoor scenes.

\subsection{Training Objective}
We introduce the process of encoding condition \(c\) into \cref{eq7}.
Therefore, the training objective of our model can be defined as follows:

\vspace{-0.5cm}
\begin{equation}\label{eq14}
\small
L(\theta,\rho,\omega)=\mathbb{E}_{t,X^{0},\epsilon}\|{\epsilon}-\epsilon_{{\theta}}({\sqrt{{\bar{\alpha}}_{t}}}{{X}}^{0}+{\sqrt{1-{\bar{\alpha}}_{t}}}{{\epsilon}},f_{\omega}(\Phi_{\rho}),t)\|^{2}.
\end{equation}
\vspace{-0.5cm}

By minimizing this loss, we can simultaneously train the encoder \(\Phi_{\rho}\), the CANet \(f_{\omega}\) and the CPDM \(\epsilon_{\boldsymbol{\theta}}\).
Intuitively, the training process encourages the encoder to extract hierarchical geometric features from the original point cloud and encourages the CPDM to reconstruct the original point cloud according to the hierarchical geometric features. In this process, the CPDM performs a task similar to point cloud completion.

\begin{table}[t]\small
\setlength{\tabcolsep}{2.0pt}
\captionsetup{skip=1.0pt}
 \caption{\textbf{Object classification results on ScanObjectNN.} We report the Overall Accuracy(\%).}
  \label{tab:obj_cls}
  \centering
  \begin{tabular}{lcccc}
    \toprule
    Methods     & Pre. & OBJ-ONLY    & OBJ-BG      & PB-T50-RS \\
    \midrule
    PointNet~\cite{qi2017pointnet} & \ding{56} & 79.2  & 73.3  & 68.0 \\
    PointNet++~\cite{qi2017pointnet++} & \ding{56} & 84.3  & 82.3  & 77.9 \\
    PointCNN~\cite{li2018pointcnn} & \ding{56} & 85.5  & 86.1  & 78.5 \\
    DGCNN~\cite{wang2019dynamic} & \ding{56} & 86.2  & 82.8  & 78.1 \\
    \midrule
    Transformer~\cite{Point-BERT}& \ding{56} & 80.55 & 79.86 & 77.24 \\
    Transformer-OcCo~\cite{Point-BERT} & \ding{56} & 85.54 & 84.85 & 78.79 \\
    Point-BERT~\cite{Point-BERT} &\ding{52} & 88.12 & 87.43 & 83.07 \\
    MaskPoint~\cite{liu2022masked} &\ding{52} & 89.70  & 89.30  & 84.60 \\
    Point-MAE~\cite{pang2022masked} &\ding{52} & 88.29 & 90.02 & 85.18 \\
    TAP~\cite{wang2023take} &\ding{52} & 89.50 & 90.36  & 85.67 \\
    \rowcolor{gray!30}
    \textbf{\ourmethod (Ours)} &\ding{52} & \textbf{91.91} & \textbf{93.29} & \textbf{87.61} \\
    \bottomrule
  \end{tabular}
  \vspace{-0.4cm}
\end{table}

\noindent\textbf{Recurrent Uniform Sampling Strategy.} 
According to \cref{eq14}, we need to sample a time step \(t\) randomly from the range [1, T] for each point cloud data for network training.
However, we observe that networks trained with samples from different time steps exhibit varying performance on downstream tasks. As illustrated in \cref{tab:time_intervals}, the encoder trained by sampling \(t\) from the early interval is more suitable for the classification task. In contrast, the encoder trained by sampling from the later interval performs better on the segmentation task.  Based on this discovery, We propose a more effective recurrent uniform sampling strategy. 
Specifically, we divide the time step range [1, T] into \(h\) intervals: \(\left\{\left[d{\times} i{+}1, d{\times}(i{+}1)\right]\right\}_{i=0}^{h-1}\) where  \(d {=} \lfloor T {/} {h} \rfloor\). As in \cref{eq15}, we randomly sample \(t\) from these \(h\) intervals for each sample data, calculate the loss \(h\) times, and average them to obtain the final loss.

\vspace{-0.5cm}
\begin{equation}\label{eq15}
\small
\mathcal{L(\theta,\rho,\omega)}=\frac{1}{h}\sum_{i=0}^{h-1}L(\theta,\rho,\omega)_{t\sim Q_i}, \quad
Q_i=[d{\times} i{+}1, d{\times}(i{+}1)].
\end{equation}
\vspace{-0.4cm}

Intuitively, this sampling strategy allows the encoder to learn different levels of geometric prior and learn from balanced supervision. It is more uniform compared to randomly sampling a single \(t\) from \([1, T]\) in the original DDPM~\cite{ho2020denoising}. Our approach divides the time steps into \(h\) = 4 intervals, as discussed in \cref{sec:ablation}.

\noindent\textbf{Discussion.} We chose to pre-train the backbone instead of the diffusion model \(\epsilon_{\boldsymbol{\theta}}\) for two reasons. Firstly, the backbone can be various deep feature extraction networks, which is more effective in extracting low-level and high-level geometric features compared to the typically simpler diffusion model \(\epsilon_{\boldsymbol{\theta}}\). Secondly, separating the backbone from the pipeline makes our pre-trained framework more adaptable to different architectures, thereby increasing its flexibility.

\begin{table}\small
\setlength{\tabcolsep}{6.5pt}
\captionsetup{skip=1.0pt}
  \caption{\textbf{Object detection results on  ScanNet.} We report the Average Precision(\%). "Pre Dataset" refers to the pre-training dataset, ScanNet-vid and ScanNet-Medium are both subsets of ScanNet.}
  \label{tab:obj_det}
  \centering
  \begin{tabular}{lccc}
    \toprule
    Methods     &Pre.     & Pre Dataset     & \(\rm AP_{50}\) \\
     \midrule
    VoteNet~\cite{qi2019deep} & \ding{56} & -     & 33.5 \\
    STRL~\cite{huang2021spatio}  & \ding{52}     & ScanNet~\cite{dai2017scannet}    & 38.4 \\
    PointContrast~\cite{xie2020pointcontrast} & \ding{52}     &  ScanNet~\cite{dai2017scannet}   & 38.0 \\
    DepthContrast~\cite{zhang2021self} & \ding{52}     & ScanNet-vid~\cite{zhang2021self} & 42.9 \\
    \midrule
    3DETR~\cite{misra2021end} & \ding{56} & -     & 37.9 \\
    Point-BERT~\cite{Point-BERT} & \ding{52}     & ScanNet-Medium~\cite{liu2022masked}  & 38.3 \\
    MaskPoint~\cite{liu2022masked} & \ding{52}     & ScanNet-Medium~\cite{liu2022masked}   & 42.1 \\
    Point-MAE~\cite{pang2022masked} & \ding{52}     & ShapeNet~\cite{chang2015shapenet}   & 42.8 \\ 
    TAP \cite{wang2023take}  & \ding{52} & ShapeNet~\cite{chang2015shapenet}  & 41.4 \\
    \rowcolor{gray!30}
    \textbf{\ourmethod (Ours)} & \textbf{\ding{52}} & ShapeNet~\cite{chang2015shapenet} & \textbf{43.7} \\
    \bottomrule
  \end{tabular}
  \vspace{-0.3cm}
\end{table}
\section{Experiments}
\subsection{Pre-training Setups} \label{sec:pretrain_setup}
\noindent\textbf{Pre-training.} We use ShapeNet~\cite{chang2015shapenet} to pre-train the model, a synthetic 3D dataset that contains 52,470 3D shapes across 55 object categories. We pre-train our model only on the training set, which consists of 41,952 shapes. For each 3D shape, we sample 1,024 points to serve as the input for the model. We set \(s\) as 64, which means each point cloud is divided into 64 patches. Furthermore, the KNN algorithm is used to select \(k\)=32 nearest points as a point patch.

\noindent\textbf{Model Configurations.}  Following ~\cite{Point-BERT, pang2022masked}, we set the embedding dimension of the transformer encoder to 384 and the number of heads to 6. The condition dimension is 768. 

\noindent\textbf{Training Details.} During pre-training, we adopt the AdamW optimizer with a weight decay of 0.05 and a learning rate of 0.001. We apply the cosine decay schedule to adjust the learning rate. Random scaling and translation are used for data augmentation. Our model is pre-trained for 300 epochs with a batchsize of 128. The \(T\) for the diffusion process is set to 2000, and \(\beta_{t}\) linearly increases from 1e-4 to 1e-2.

\noindent\textbf{Visualization.} To demonstrate the effectiveness of our pre-training scheme, we visualize the point cloud generated by our \ourmethod.
As shown in \cref{fig:visualization}, we apply a high mask ratio of 0.8 to the input point cloud for masking and use the masked point cloud as a condition to guide the diffusion model in generating the original point cloud. Our \ourmethod produces high-quality point clouds. Experimental results demonstrate that the geometric prior learned through our pre-training method can provide excellent guidance for both shallow texture and shape semantics.

\begin{table}\small
\setlength{\tabcolsep}{9.5pt}
\captionsetup{skip=1.0pt}
  \caption{\textbf{Semantic segmentation results on S3DIS Area 5.} We report the mean IoU(\%) and mean Accuracy(\%).}
  \label{tab:indoor_seg}
  \centering
  \begin{tabular}{lccc}
    \toprule
    Methods     &Pre.     & mIoU    & mAcc \\
     \midrule
    PointNet~\cite{qi2017pointnet} & \ding{56}     & 41.1  & 49.0 \\
    PointNet++~\cite{qi2017pointnet++} & \ding{56}  & 53.5  &-  \\
    PointCNN~\cite{li2018pointcnn}  & \ding{56}  & 57.3  & 63.9  \\
    KPConv~\cite{thomas2019kpconv} & \ding{56}  & 67.1  & 72.8 \\
    SegGCN~\cite{lei2020seggcn} & \ding{56}  & 63.6  & 70.4 \\
    Pix4Point~\cite{qian2022pix4point} & \ding{56}     & 69.6  & 75.2 \\
    MKConv~\cite{woo2023mkconv} & \ding{56}     & 67.7  & 75.1 \\
    \midrule
    PointNeXt~\cite{qian2022pointnext} & \ding{56}     & 68.5 & 75.1 \\
    Point-BERT~\cite{Point-BERT} & \ding{52}     & 68.9 & 76.1 \\ 
    MaskPoint~\cite{liu2022masked}& \ding{52}     & 68.6 & 74.2 \\ 
    Point-MAE~\cite{pang2022masked}& \ding{52}     & 68.4 & 76.2\\ \rowcolor{gray!30}
    \textbf{\ourmethod (Ours)} & \textbf{\ding{52}} & \textbf{70.0} & \textbf{77.1} \\
    \bottomrule
  \end{tabular}
  \vspace{-0.41cm}
\end{table}

\subsection{Downstream Tasks}
A high-quality point cloud pre-trained model should perceive hierarchical geometric prior. To assess the efficacy of the pre-trained model, we gauged its performance on various fine-tuned tasks using numerous real-world datasets.

\begin{table*} \small 
\setlength{\tabcolsep}{2.0pt}
\captionsetup{skip=1.2pt}
  \caption{\textbf{Semantic segmentation results on SemanticKITTI val set.}
  We report the mean IoU(\%) and IoU(\%) for some semantic classes. 
}
  \centering
  \label{tab:outdoor_seg}
  \begin{tabular}{l|c|ccccccccccccc}
    \toprule
    
    Methods & mIoU & car & bicycle & truck &preson & bicyclist &motorcyclist &road & sidewalk & parking & vegetation &trunk & terrain \\
     \midrule
    Cylinder3D~\cite{zhu2021cylindrical} & 66.1  & 96.9  & 54.4  & 81.0 & 79.3   & 92.4  & 0.1 & 94.6 & 82.2   & 47.9  & 85.9  & 66.9  & 69.2 \\
    SPVCNN~\cite{tang2020searching} & 68.6  & 97.9 & 59.8 & 79.8 & 80.0 & 92.0  & 0.6   & 94.2   & 81.7   & 50.4  & 88.0    & 69.7 & 74.1 \\
    RPVNet~\cite{xu2021rpvnet} & 68.9  & 97.9 & 42.8  & 91.2 & 78.3 & 90.2 & 0.7   & 95.2 & 83.1   & 57.1 & 87.3  & 71.4  & 72.0 \\
     \midrule
    MinkowskiNet~\cite{choy20194d} & 70.2  & 97.4  & 56.1  & 84.0  &\textbf{81.9}  & 91.4 & 24.0  & 94.0 & 81.3    & 52.2  & 88.4  & 68.6  & 74.8 \\ \rowcolor{gray!30}
    \textbf{MinkowskiNet+\ourmethod} & \textbf{71.3} & \textbf{97.5}  & \textbf{58.8}  & \textbf{92.8} &81.4 & \textbf{92.3} & \textbf{30.3} & \textbf{94.1} & \textbf{81.7} & \textbf{56.0}    & \textbf{88.5} & \textbf{69.1}  & \textbf{75.2} \\ 
    \bottomrule
  \end{tabular}
  \vspace{-0.2cm}
\end{table*}

\begin{table}\small
\vspace{-0.2cm}
  \centering
  \setlength{\tabcolsep}{20.0pt}
  \captionsetup{skip=1.2pt}
  \caption{\textbf{Object detection results of CAGroup3D with and without pre-training.} 
  We report the Average Precision(\%). }
  \label{tab:CAGroup3D_pretrain}
    \begin{tabular}{lrr}
    \toprule
    Methods & \(\rm AP_{25}\)  &  \(\rm AP_{50}\)  \\
    \midrule
    CAGroup~\cite{wang2022cagroup3d} & 73.20  & 60.84 \\ \rowcolor{gray!30}
    \textbf{CAGroup+PointDif} & \textbf{74.14} & \textbf{61.31} \\
    \bottomrule
    \end{tabular}
  \vspace{-0.1cm}
\end{table}%

\noindent\textbf{Object Classification.} We first use the classification task on ScanObjectNN~\cite{uy2019revisiting} to evaluate the shape recognition ability of the pre-trained model by \ourmethod.  The ScanObjectNN dataset is divided into three subsets: OBJ-ONLY (only objects), OBJ-BG (objects and background), and PB-T50-RS (objects, background, and artificially added perturbations). We take the Overall Accuracy on these three subsets as the evaluation metric, and the detailed experimental results are summarized in \cref{tab:obj_cls}. Our \ourmethod achieves better performance on all subsets, exceeding TAP by 2.4\(\%\), 2.9\(\%\) and 1.9\(\%\), respectively. The significant improvement on the challenging ScanObjectNN benchmark strongly validates the effectiveness of our model in shaping understanding.

\noindent\textbf{Object Detection.} We validate our model on the more challenging indoor dataset ScanNetV2~\cite{dai2017scannet} for 3D object detection task to assess the  scene understanding ability. 
We adopt 3DETR~\cite{misra2021end} as our method's task head. To ensure a fair comparison, we follow MaskPoint~\cite{liu2022masked} and replace the encoder of 3DETR with our pre-trained encoder and fine-tune it. Unlike MaskPoint and Point-BERT, which are pre-trained on the ScanNet-Medium dataset in the same domain as ScanNetV2, our approach and Point-MAE are pre-trained on ShapeNet in a different domain and only fine-tuned on the training set of ScanNetV2. \cref{tab:obj_det} displays our experimental results. Our method outperforms Point-MAE and surpasses MaskPoint and Point-BERT by 1.6\% and 5.4\%, respectively. Additionally, our approach exhibits a 2.3\% improvement compared to pre-training the transformer encoder of 3DETR on the ShapeNet dataset using the TAP method. The experiments demonstrate that our model exhibits strong transferability and generalization capability on scene understanding.

\noindent\textbf{Indoor Semantic Segmentation.} We further validate our model on the indoor S3DIS dataset~\cite{armeni20163d} for semantic segmentation tasks to show the understanding of contextual semantics and local geometric relationships. We test our model on Area 5 while training on other areas. To make a fair comparison, we put all pre-trained models in the same codebase based on the PointNext~\cite{qian2022pointnext} baseline and use the same decoder and semantic segmentation head. We freeze the encoder pre-trained on  ShapeNet and fine-tune the decoder and the segmentation head. The experiment results are shown in \cref{tab:indoor_seg}. Compared to training from scratch, our method boosts the performance of PointNext by 1.5\% in terms of mIoU. Compared to other pre-training methods such as Point-BERT, MaskPoint and Point-MAE, our method achieves approximately 1.4\% improvement for each on mIoU. Note that,  PointNext is originally trained using a batchsize of 8, since computational resource constraints, we thus retrained it with a batchsize of 4 for a fair comparison. Significant improvements indicate that our pre-trained model has successfully acquired hierarchical geometric prior knowledge essential for comprehending contextual semantics and local geometric relationships.

\begin{table}\small
\setlength{\tabcolsep}{16.0pt}
\captionsetup{skip=1.2pt}
  \caption{\textbf{Conditional guidance strategies.} We report the mean IoU(\%) and mean Accuracy(\%) on S3DIS Area 5.}
  \label{tab:cond_strategy}
  \centering
  \begin{tabular}{lccc}
    \toprule
    Methods     &mIoU      & mAcc \\
     \midrule
    Cross Attention & 69.09  & 75.19 \\
    Point Concat & 69.43 & 75.39 \\
     \rowcolor{gray!30}
    \textbf{Point Condition Network} & \textbf{70.02} & \textbf{77.05} \\
    \bottomrule
  \end{tabular}
  \vspace{-0.4cm}
\end{table}

\noindent\textbf{Outdoor Semantic Segmentation.} We also validate the effectiveness of our method on the more challenging real-world outdoor scene dataset KITTI. The SemanticKITTI dataset~\cite{behley2019semantickitti} is a large-scale outdoor LiDAR segmentation dataset, consisting of 43,000 scans with 19 semantic categories. We employ MinkowskiNet~\cite{choy20194d} as our baseline model. During the pre-training phase, we discard its segmentation head and utilize the backbone MinkUNet as the encoder to extract latent features. We pre-train the MinkUNet using our framework on ShapeNet and subsequently fine-tuned it on the SemanticKITTI. Other pre-training configurations follow the guidelines outlined in Sec.~\ref{sec:pretrain_setup}. The experiment results in \cref{tab:outdoor_seg} demonstrate that our pre-training method achieves 71.3\% mIoU, which is a 1.1\% improvement over the train-from-scratch variant. Our pre-training framework for point-to-point guided generation can assist the backbone in learning density priors and enable it to adapt to downstream tasks with significant density variations. The entire results are reported in \cref{sec:additional_results}.

\noindent\textbf{Object detection results of CAGroup3D with and without pre-training.}
We further evaluate our pre-training method on the competitive 3D object detection model, CAGroup3D~\cite{wang2022cagroup3d}, a two-stage fully sparse 3D detection network. We train CAGroup3D from scratch and report the result for a fair comparison. We use our method to pre-train the backbone BiResNet on ShapeNet. Specifically, we treat BiResNet as the encoder to extract features. The conditional point generator employs the masked features to guide the point-to-point recovery of the original point cloud. Other pre-training settings follow \cref{sec:pretrain_setup}. The experimental results are shown in \cref{tab:CAGroup3D_pretrain}. Compared to the train-from-scratch variant, our method improves performance by 0.9\% and 0.5\% on \(\rm AP_{25}\) and \(\rm AP_{50}\), respectively. Therefore, our pre-training framework can be flexibly applied to various backbones to improve performance. Please refer to \cref{sec:additional_results} for additional results.

\begin{table} \small
\setlength{\tabcolsep}{2.4pt}
\vspace{-0.3cm}
\captionsetup{skip=1.2pt}
  \caption{\textbf{Recurrent uniform sampling.} `\#Point Clouds' represents the number of unique point clouds in a batch, and `\#$t$' represents the number of time steps \(t\) sampled for each point cloud.
}
  \centering
  \label{tab:sampling_strategy}
  \begin{tabular}{cccccc}
    \toprule
    \#Point Clouds & \#\(t\) & Intervals & Effective Batchsize & mIoU    &mAcc \\
     \midrule
     \rowcolor{gray!30}
    \textbf{128} & \textbf{4} & \textbf{4} & 512 &\textbf{70.02}  & \textbf{77.05} \\
    128   & 4     & 1 & 512   & 69.68  & 75.90 \\
    
    \midrule
    256   & 2     & 2   & 512  &69.67    &76.26  \\
    256   & 2     & 1  & 512   &69.36    &75.94
  \\
    \midrule
    64    & 8     & 8  & 512   & 69.42   & 75.71 \\
    64    & 8     & 1  & 512   & 69.24   & 75.50 \\
    \midrule
    512 & 1     & 4  & 512   & 69.91    & 75.93 \\
    512 & 1     & 1  & 512   & 69.51    & 75.95 \\
    \midrule
    128  &1      &1    &128   &69.39    & 76.45 \\
    128   &3      &3   & 384   & 69.63  & 75.54 \\
    128   &5      &5   & 640   & 69.24   & 75.16 \\
    \bottomrule
  \end{tabular}
  \vspace{-0.4cm}
\end{table}

\subsection{Ablation Study} 
\label{sec:ablation}
\noindent\textbf{Conditional guidance strategies.} We study the influence of different guidance strategies for CPDM on S3DIS.
As shown in \cref{tab:cond_strategy}, the cross-attention way even performs worse than the simple pointwise concatenation way. We speculate this is because the cross-attention mechanism attempts to capture relationships between different points. However, the density varies across different regions for point cloud data, potentially impacting the model's performance. In contrast, our PCNet employs a point-to-point guidance approach, where each point is processed independently of others. This approach is advantageous in enabling the network to capture point density information. Additionally, compared to pointwise concatenation, our utilization of the reset gate control mechanism assists the network in adaptively retaining relevant geometric features, thereby enhancing performance.

\noindent\textbf{Recurrent uniform sampling.} We validate the effectiveness of our proposed recurrent uniform sampling strategy on S3DIS. Specifically, 
(i) we first verify the impact of the number of partition intervals and whether the recurrent sampling strategy is adopted on experimental results with the same effective batchsize. As presented in lines 1-6 of \cref{tab:sampling_strategy}, each pair of lines illustrates the results obtained with and without recurrent uniform sampling. The results indicate that our sampling strategy outperforms the original random sampling method under the same effective batch size. 
(ii) We further investigate the impact of sample diversity on the experimental results with the same effective batchsize. Our approach involves sampling \(t\) 4 times and calculating the loss for each sample. We increase the number of unique point clouds in a batch by a factor of 4, which is equivalent to sampling only one \(t\) for each point cloud sample. For the experiment in line 7 of \cref{tab:sampling_strategy}, we uniformly sample from 4 intervals for each set of 4 adjacent samples. The experimental results further demonstrate the superiority of our recurrent uniform sampling method for each sample.
(iii) We also validate the experimental results by partitioning different numbers of intervals and performing uniform sampling, while keeping the number of unique point clouds in a batch constant. The results in lines 10-11 of \cref{tab:sampling_strategy} indicate that our algorithm, which divides the samples into 4 intervals and performs recurrent uniform sampling, is optimal. Compared to the original sampling method in DDPM (line 9 of \cref{tab:sampling_strategy}), our recurrent uniform sampling strategy resulted in a 0.6\% performance improvement.

\begin{figure}
    \centering
\includegraphics[width=0.45\textwidth]{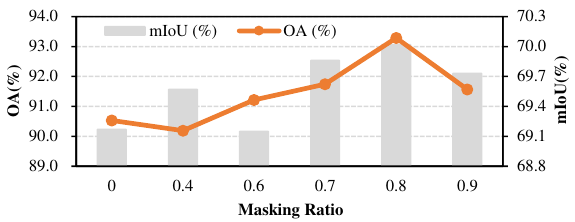}
    \captionsetup{skip=2.0pt}
    \caption{\textbf{Masking ratio.} We report the Overall Accuracy(\%) on ScanObjectNN and the mean IoU(\%) on S3DIS with different masking ratios.
}
    \label{fig:masking}
    \vspace{-0.5cm}
\end{figure}

\begin{table} \small
\setlength{\tabcolsep}{1.3pt}
\captionsetup{skip=1.2pt}
  \caption{\textbf{Different time intervals.} We study the impact of pre-training with different time intervals. 
  We report the object classification results on ScanObjectNN and semantic segmentation results on S3DIS Area 5.
}
  \centering
  \label{tab:time_intervals}
  \begin{tabular}{l|ccc|c}
    \toprule
          \multirow{2}{*}{Time Intervals} & \multicolumn{3}{c|}{Classification} & \multicolumn{1}{c}{Segmentation} \\
          \specialrule{0em}{1pt}{1pt}
& OBJ-ONLY & OBJ-BG & PB-T50-RS & mIoU \\
    \midrule
    \(\rm[1, 500]\) & \textbf{92.43} & 92.25 & \textbf{88.31} & 68.83\\
    \(\rm[501, 1000]\) & 91.57 & 91.39 & 87.23 & 68.52\\
    \(\rm[1001, 1500]\) & 90.36 & 92.25 & 87.13 & 69.19\\
    \(\rm[1501, 2000]\) & 89.50  & 87.61 & 83.28 & 69.70 \\ \rowcolor{gray!30}
    \textbf{\(\rm[1, 2000]\)(Ours)} & 91.91 & \textbf{93.29} & 87.61 & \textbf{70.02} \\
    \bottomrule
  \end{tabular}
  \vspace{-0.6cm}
\end{table}

\noindent\textbf{Different time intervals.} To demonstrate that our pre-training method learns hierarchical geometric prior, we conduct experiments with the same settings by sampling $t$ at different intervals for pre-training and evaluating the results. \cref{tab:time_intervals} shows that the classification results are significantly better in the \(\rm[1, 500]\) time interval than in other intervals, while achieving unsatisfactory segmentation results. Conversely, the segmentation performance is better in the \(\rm[1501, 2000]\) time interval, while the classification results will be slightly worse. We observe a gradual transition of classification and segmentation results among these four intervals, which fully validates our theory. In the early intervals of training, the model needs more low-level geometric features to guide the recovery of shallow texture from low-noise point clouds. Moreover, in the later intervals, high-level geometric features become crucial for guiding the recovery of semantic structure in high-noise point clouds. Therefore, our model can learn hierarchical geometric features throughout the entire training process.

\noindent\textbf{Masking ratio.}
We further validate the impact of different masking ratios on downstream tasks and separately report the results for classification on ScanObjectNN and semantic segmentation on S3DIS. As shown in \cref{fig:masking}, encoding all point patches without masking harms the model's learning. By employing masking, the overall difficulty of the self-supervised proxy task is increased, thereby aiding the backbone in learning more rich geometric priors. Additionally, our method achieves the best classification and semantic segmentation performance when the mask ratio is 0.8.
\section{Conclusion}

In conclusion, we propose a novel framework for point cloud pre-training based on diffusion models, called \ourmethod. It enables the backbone to learn hierarchical geometric prior through the progressive guidance characteristic of the conditional diffusion model. Specifically, we present a conditional point generator to assist the network in learning the point density distribution of the object through point-to-point guidance generation. We also introduce a recurrent uniform sampling strategy on time steps to facilitate the balanced supervision during the backbone's pre-training. Our extensive experiments on various real-world indoor and outdoor datasets demonstrate significant performance improvements compared to existing methods. Moreover, our proposed method consistently increases performance on competitive backbones. Overall, our diffusion-based pre-training framework provides a new direction for advancing point cloud processing.
{
    \small
    \bibliographystyle{ieeenat_fullname}
    \bibliography{main}
}

\clearpage
\maketitlesupplementary

\section{Proof}
\label{sec:proof}
Calculating the probability distribution \(q(X^{t-1}|X^{t})\) for the reverse process is hard. However, given \(X^{0}\), the posterior of the forward diffusion process can be calculated using the following equation:

\vspace{-0.05cm}
\begin{equation}
\small
q(X^{t-1}|X^{t},X^{0}) {=} N(X^{t-1};\tilde{\mu}_{t}(X^{t},X^{0}), \tilde{\beta_{t}}I), \
\end{equation}
\vspace{-0.4cm}

\vspace{-0.5cm}
\begin{equation}\label{eq17}
\small
\tilde{\mu}_{t}(X^{t},X^{0}) {=} \frac{1}{\sqrt{\alpha_{t}}}(X^{t} {-} \frac{\beta_{t}}{\sqrt{1-\bar{\alpha}_{t}}} \epsilon), \
\tilde{\beta_{t}} {=} \frac{\beta_{t}(1-\bar{\alpha}_{t-1})}{1-\bar{\alpha}_{t}}.
\end{equation}
\vspace{-0.3cm}

According to \cref{eq6} in the main text, the variational lower bound can be divided into three parts:

\vspace{-0.3cm}
\begin{equation}
\small
\begin{split}
&E_{q}[\underbrace{-log p_{\theta}(X^{0}|X^{1},c)}_{L_0} + \underbrace{D_{KL}(q(X^{T}|X^{0})||p(X^{T}))]}_{L_{T}} \\ &+ \sum\limits_{t=2}\limits^{T}\underbrace{D_{KL}(q(X^{t-1}|X^{t},X^{0})||p_{\theta}(X^{t-1}|X^{t},c))}_{L_{t-1}}.
\end{split}
\end{equation}
\vspace{-0.4cm}

\noindent \(L_{T}\) is a constant without parameters and can be ignored. To compute the parameterization of \(L_{t-1}\), following~\cite{ho2020denoising}, we set the mean \(\mu_{\theta}(X^{t},t,c)\) of \(p_{\theta}(X^{t-1}|X^{t},c)\) to:

\vspace{-0.4cm}
\begin{equation}\label{eq19}
\small
\mu_{\theta}(X^{t},c,t) {=} \frac{1}{\sqrt{\alpha_{t}}}(X^{t} {-} \frac{\beta_{t}}{\sqrt{1-\bar{\alpha}_{t}}} \epsilon_{\theta}(X^{t},c,t)).
\end{equation}

\noindent We can calculate \(L_{t-1}\):
\vspace{-0.2cm}
\begin{equation}
\small
\begin{split}
L_{t-1}=\mathbb{E}_{q}\left[\frac{1}{2\sigma_{t}^{2}}||\tilde{\mu}_{t}(X^{t},X^{0})-\mu_{\theta}(X^{t},c,t)||^{2}\right]+C,
\end{split}
\vspace{-0.2cm}
\end{equation}

\noindent where \(C\) is a parameter-free constant that can be disregarded. By substituting \cref{eq17} and \cref{eq19} into \(L_{t-1}\):

\vspace{-0.2cm}
\begin{equation}
\small
\begin{split}
&L_{t-1}= \mathbb{E}_{t,X^{t}, c, \epsilon}\left[\frac{1}{2\sigma_{t}^{2}}||\frac{1}{\sqrt{\alpha_{t}}}(X^{t} {-} \frac{\beta_{t}}{\sqrt{1-\bar{\alpha}_{t}}} \epsilon) \right. \\
&\left. \text{\quad\quad\quad\quad\quad} -\frac{1}{\sqrt{\alpha_{t}}}(X^{t}{-} \frac{\beta_{t}}{\sqrt{1-\bar{\alpha}_{t}}} \epsilon_{\theta}(X^{t},c,t))||^{2}\right] \\
&=\mathbb{E}_{t,X^{t}, c,\epsilon}\left[\frac{\beta_{t}^{2}}{2\sigma_{t}^{2}\alpha_{t}(1-\bar{\alpha}t)}||\epsilon-\epsilon_{\theta}(X^{t},c,t)||^{2}\right] \\
&=\mathbb{E}_{t,X^{0}, c, \epsilon}\left[\frac{\beta_{t}^{2}}{2\sigma_{t}^{2}\alpha_{t}(1-\bar{\alpha}t)}||\epsilon-\epsilon_{\theta}({\sqrt{{\bar{\alpha}}_{t}}}{{X}}^{0}+{\sqrt{1-{\bar{\alpha}}_{t}}}{\epsilon},c,t)||^{2}\right],
\end{split}
\vspace{-0.2cm}
\end{equation}

\noindent where \(\frac{\beta_{t}^{2}}{2\sigma_{t}^{2}\alpha_{t}(1-\bar{\alpha}t)}\) is a constant that is unrelated to the loss, and following ~\cite{ho2020denoising}, we can further simplify the training loss:

\vspace{-0.1cm}
\begin{equation}
\small
L({\theta})=\mathbb{E}_{t,X^{0}, c, \epsilon} \left[\|{\epsilon}-\epsilon_{{\theta}}({\sqrt{{\bar{\alpha}}_{t}}}{{X}}^{0}+{\sqrt{1-{\bar{\alpha}}_{t}}}{{\epsilon}},c,t)\|^{2}\right].
\end{equation}
\vspace{-0.6cm}

\section{Implementation Details}
All experiments are conducted on the RTX 3090 GPU. We describe the details of fine-tuning on various tasks.

\noindent\textbf{Object classification.} We use a three-layer MLP with dropout as the classification head. During the fine-tuning process, we sample 2048 points for each point cloud, divide them into 128 point patches, set the learning rate to 5e-4, and fine-tune for 300 epochs.

\noindent\textbf{3D object detection.} Unlike MaskPoint~\cite{liu2022masked}, which is pre-trained on ScanNet-Medium and loads the weights of both the SA layer and the encoder during fine-tuning. During the fine-tuning stage, we only load the weights of the transformer encoder pre-trained on ShapeNet~\cite{chang2015shapenet}. Following Maskpoint, we set the learning rate to 5e-4 and use the AdamW optimizer with a weight decay of 0.1. The batch size is set to 8.

\begin{table*} \footnotesize
\setlength{\tabcolsep}{3.0pt}
\captionsetup{skip=1.5pt}
  \caption{\textbf{Semantic segmentation results on SemanticKITTI val set.} We report the mean IoU(\%) and the IoU(\%) for all semantic classes.
}
  \centering
  \label{tab:outdoor_seg_all}
  \begin{tabular}{lr|rrrrrrrrrrrrrrrrrrr}
    \toprule
    Methods & \multicolumn{1}{c|}{\begin{sideways}mIoU\end{sideways}} & \multicolumn{1}{c}{\begin{sideways}car\end{sideways}} & \multicolumn{1}{c}{\begin{sideways}bicycle\end{sideways}} & \multicolumn{1}{c}{\begin{sideways}motorcycle\end{sideways}} & \multicolumn{1}{c}{\begin{sideways}truck\end{sideways}} & \multicolumn{1}{c}{\begin{sideways}other-vehicle\end{sideways}} & \multicolumn{1}{c}{\begin{sideways}person\end{sideways}} & \multicolumn{1}{c}{\begin{sideways}bicyclist\end{sideways}} & \multicolumn{1}{c}{\begin{sideways}motorcyclist\end{sideways}} & \multicolumn{1}{c}{\begin{sideways}road\end{sideways}} & \multicolumn{1}{c}{\begin{sideways}parking\end{sideways}} & \multicolumn{1}{c}{\begin{sideways}sidewalk\end{sideways}} & \multicolumn{1}{c}{\begin{sideways}other-ground\end{sideways}} & \multicolumn{1}{c}{\begin{sideways}building\end{sideways}} & \multicolumn{1}{c}{\begin{sideways}fence\end{sideways}} & \multicolumn{1}{c}{\begin{sideways}vegetation\end{sideways}} & \multicolumn{1}{c}{\begin{sideways}trunk\end{sideways}} & \multicolumn{1}{c}{\begin{sideways}terrain\end{sideways}} & \multicolumn{1}{c}{\begin{sideways}pole\end{sideways}} & \multicolumn{1}{c}{\begin{sideways}traffic-sign\end{sideways}} \\
    \midrule
    Cylinder3D~\cite{zhu2021cylindrical} & 66.1  & 96.9  & 54.4  & 75.9  & 81.0  & 67.0  & 79.3  & 92.4  & 0.1   & 94.6  & 47.9  & 82.2  & 0.1   & 90.3  & 57.0  & 85.9  & 66.9  & 69.2  & 63.6  & 50.6  \\
    SPVCNN~\cite{tang2020searching} & 68.5  & 97.9  & 59.8  & 81.1  & 79.8  & 80.8  & 80.0  & 92.0  & 0.6   & 94.2  & 50.4  & 81.7  & 0.6   & 90.9  & 63.5  & 88.0  & 69.7  & 74.1  & 65.8  & 51.5  \\
    RPVNet~\cite{xu2021rpvnet} & 68.9  & 97.9  & 42.8  & 87.6  & 91.2  & 83.5  & 78.3  & 90.2  & 0.7   & 95.2  & 57.1  & 83.1  & 0.2   & 91.0  & 63.2  & 87.3  & 71.4  & 72.0  & 64.9  & 51.5  \\
    \midrule
    MinkowskiNet~\cite{choy20194d} & 70.2  & 97.4  & 56.1  & \textbf{84.9} & 84.0  & 79.1  & \textbf{81.9} & 91.4  & 24.0  & 94.0  & 52.2  & 81.3  & 0.2   & \textbf{92.0} & \textbf{67.2} & 88.4  & 68.6  & 74.8  & \textbf{65.5} & \textbf{50.6}  \\ \rowcolor{gray!30}
    \textbf{MinkowskiNet+PointDif} & \textbf{71.3} & \textbf{97.5} & \textbf{58.8} & 84.6  & \textbf{92.8} & \textbf{80.6} & 81.4  & \textbf{92.3} & \textbf{30.3} & \textbf{94.1} & \textbf{56.0} & \textbf{81.7} & \textbf{0.2} & 91.4  & 65.4  & \textbf{88.5} & \textbf{69.1} & \textbf{75.2} & 65.0  & 50.5  \\
    \bottomrule
  \end{tabular}
\end{table*}

\begin{table*} \footnotesize
\setlength{\tabcolsep}{1.4pt}
\captionsetup{skip=1.2pt}
  \caption{\textbf{Object detection results of CAGroup3D with and without pre-training.} We report the Overall and different category results at \(\rm AP_{25}\)(\%) and \(\rm AP_{50}\)(\%). 
}

  \centering
  \label{tab:CAGroup3D_pretrain_all}
  \begin{tabular}{ccc|cccccccccccccccccc}
    \toprule
          Methods & \begin{sideways}Metric\end{sideways} & \begin{sideways}Overall\end{sideways} & \begin{sideways}cabinet\end{sideways} & \begin{sideways}bed\end{sideways} & \begin{sideways}chair\end{sideways} & \begin{sideways}sofa\end{sideways} & \begin{sideways}table\end{sideways} & \begin{sideways}door\end{sideways} & \begin{sideways}window\end{sideways} & \begin{sideways}bookshelf\end{sideways} & \begin{sideways}picture\end{sideways} & \begin{sideways}counter\end{sideways} & \begin{sideways}desk\end{sideways} & \begin{sideways}curtain\end{sideways} & \begin{sideways} refrigerator\end{sideways} & \begin{sideways}showercurtrain\end{sideways} & \begin{sideways}toilet \end{sideways} & \begin{sideways}sink\end{sideways} & \begin{sideways}bathtub\end{sideways} & \begin{sideways}garbagebin\end{sideways} \\
    \midrule
    CAGroup3D~\cite{wang2022cagroup3d} & \(\rm AP_{25}\)  & 73.20  & \textbf{54.39}  & 85.78  & \textbf{95.70}  & \textbf{91.95}  & 69.67  & 67.87  & \textbf{60.84}  & 63.71  & 38.70  & 73.62  & 82.12  & 66.96  & \textbf{58.32}  & 75.80  & \textbf{99.97}  & 77.85  & 87.74  & 66.61  \\ \rowcolor{gray!30}
    \textbf{CAGroup3D+PointDif} & \textbf{\(\rm AP_{25}\)} & \textbf{74.14 } & 53.71  & \textbf{87.85} & 95.46  & 89.73  & \textbf{73.01} & \textbf{69.36} & 59.72  & \textbf{65.22} & \textbf{41.65} & \textbf{75.07} & \textbf{82.66} & \textbf{67.10} & 56.27  & \textbf{79.22} & 99.91  & \textbf{82.27} & \textbf{89.55} & \textbf{66.69} \\
    \midrule
    CAGroup3D~\cite{wang2022cagroup3d} & \(\rm AP_{50}\)  & 60.84  & \textbf{39.01}  & 81.51  & 90.24  & \textbf{82.75}  & 65.89  & 53.47  & \textbf{36.39}  & 55.82  & 25.13  & \textbf{42.01}  & \textbf{66.19}  & \textbf{49.33}  & \textbf{53.16}  & 57.73  & \textbf{96.52}  & \textbf{53.80}  & \textbf{86.75}  & \textbf{59.35}  \\ \rowcolor{gray!30}
    \textbf{CAGroup3D+PointDif} & \textbf{\(\rm AP_{50}\)} & \textbf{61.31} & 38.47  & \textbf{82.46} & \textbf{91.03} & 82.23  & \textbf{67.09} & \textbf{53.88} & 34.72  & \textbf{56.80} & \textbf{31.34} & 40.02  & 65.49  & 48.19  & 51.40  & \textbf{70.57} & 96.37  & 52.60  & 82.33  & 58.53  \\
    \bottomrule
  \end{tabular}
  \vspace{-0.3cm}
\end{table*}

\noindent\textbf{Semantic segmentation on indoor dataset.} For a fair comparison, we put all pre-trained transformer encoders within the same codebase and freeze them while fine-tuning the decoder and semantic segmentation head. Due to limited computing resources, we set the batch size to 4 during fine-tuning. The remaining settings followed those used for training PointNeXt~\cite{qian2022pointnext} from scratch in the original paper.

\noindent\textbf{Semantic segmentation on outdoor dataset.} During fine-tuning, we load the backbone MinkUNet pre-trained on ShapeNet. And fine-tune the entire network while following the same settings used for training MinkowskiNet~\cite{choy20194d} from scratch.

\noindent\textbf{3D object detection of CAGroup3D with and without pre-training.} We load the weights of the backbone BiResNet, which is pre-trained on ShapeNet using our method. Then, we fine-tune the entire CAGroup3D~\cite{wang2022cagroup3d} model using the same settings as those used for training CAGroup3D from scratch. Note that, we utilize the official codebase of CAGroup3D and consider the best-reproduced results as the baseline for comparison.

\section{Additional results}
\label{sec:additional_results}
\noindent\textbf{Semantic segmentation on outdoor dataset.} As shown in \cref{tab:outdoor_seg_all}, We report the mean IoU(\%) and the IoU(\%) on SemanticKITTI~\cite{behley2019semantickitti} for all semantic classes for different methods. Our method improves mean IoU and IoU for multiple categories compared to the variant trained from scratch. The experimental results also demonstrate that our method performs well on outdoor datasets.

\noindent\textbf{Object detection results of CAGroup3D with and without pre-training.} We report the Overall and different category results at \(\rm AP_{25}\)(\%) and \(\rm AP_{50}\)(\%). From \cref{tab:CAGroup3D_pretrain_all}, we observe that pre-training with our method leads to better performance than training CAGroup3D from scratch. Therefore, our pre-training framework can be flexibly applied to various backbones to improve performance.

\noindent\textbf{Recurrent uniform sampling.} Keeping the number of unique point clouds in a batch constant, we conduct experiments with 2 and 8 intervals divisions. The results are shown in \cref{tab:sampling_strategy_add}, our strategy of dividing the 4 intervals and uniform sampling time step \(t\) is optimal.

\noindent\textbf{Masking strategy.} We report the experimental results for downstream classification and semantic segmentation tasks with different masking strategies. The strategy of block masking involves masking adjacent point patches. From \cref{tab:masking_strategy}, we observe that random masking performs better than block masking under the same masking ratio (0.8).

\begin{table} \small 
\setlength{\tabcolsep}{2.3pt}
  \caption{\textbf{Recurrent uniform sampling.} `\#Point Clouds' represents the number of unique point clouds in a batch, and `\#$t$' represents the number of time steps \(t\) sampled for each point cloud. We report the mean IoU(\%) and mean Accuracy(\%) on S3DIS.
}
\label{tab:sampling_strategy_add}
  \centering
  \begin{tabular}{ccccccc}
    \toprule
    \#Point Clouds & \#\(t\) & Intervals & Effective Batchsize & mIoU    &mAcc \\
     \midrule
     128   &2      &2   & 256   & 69.52 & 75.46 \\ \rowcolor{gray!30}
    \textbf{128} & \textbf{4} & \textbf{4} & 512 &\textbf{70.02} & \textbf{77.05} \\
    128   &8      &8   & 1024  & 69.49 & 76.50 \\
    \bottomrule
  \end{tabular}
  \vspace{-0.3cm}
\end{table}

\begin{figure*}
\vspace{-0.6cm}
  
    \centering
    \includegraphics[width=0.99 \textwidth]{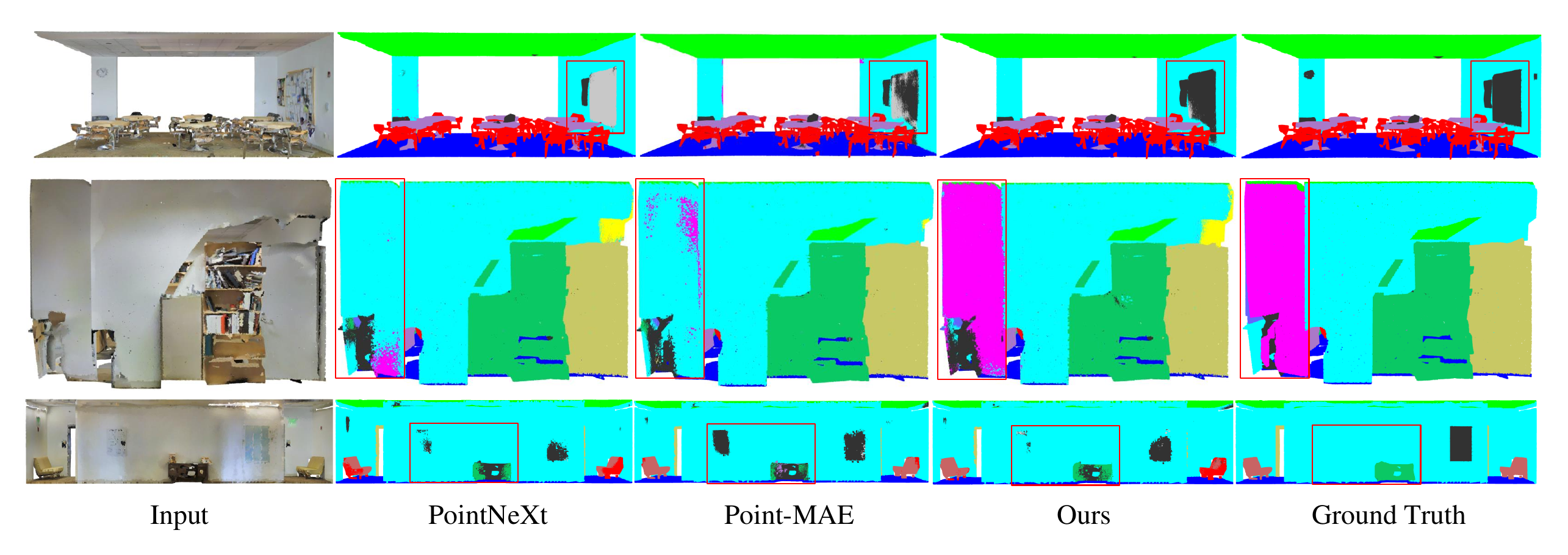}
    \captionsetup{skip=1.2pt}
    \caption{\textbf{Qualitative comparison on S3DIS semantic segmentation.} The first column shows the original point cloud input, followed by columns 2-4, which display the segmentation results of PointNeXt, Point-MAE, and our method. The fifth column shows the ground truth.}
    \label{fig:visualization_s3dis}
\end{figure*}

\section{Additional Visualization.}
\textbf{S3DIS semantic segmentation visualizations.} We provide a qualitative comparison of results for S3DIS semantic segmentation. As shown in \cref{fig:visualization_s3dis}, the predictions of our method are closer to the ground truth and less incorrectly segmented than training PointNeXt from scratch and Point-MAE.

\begin{table} \small 
\setlength{\tabcolsep}{4.0pt}
  \caption{\textbf{Masking strategy.} "Random" refers to Random masking and "Block" refers to Block masking, We report the Overall Accuracy(\%) on ScanObjectNN OBJ-BG subset  and the mean IoU(\%) on S3DIS.
}
\label{tab:masking_strategy}
  \centering
  \begin{tabular}{ccccccc}
    \toprule
    Masking Strategy & Mask Ratio & OBJ-BG & mIoU \\
     \midrule
     
     Block &0.8   &  \textbf{91.91}  & 69.47\\ \rowcolor{gray!30}
     \textbf{Random} &0.8  & \textbf{93.29} &\textbf{70.02}\\
    \bottomrule
  \end{tabular}
  \vspace{-0.3cm}
\end{table}

\section{limitation}
Our pre-training method has demonstrated outstanding performance on various 3D real datasets, but its performance is slightly worse on synthetic datasets. We suspect that this is due to the inability of synthetic datasets to fully simulate the complexity of real-world objects, such as the presence of more noise and occlusion in real datasets. Furthermore, the synthetic datasets are relatively simple, and the performance on the synthetic datasets is currently saturated, with only slight improvements from other pre-training methods. Therefore, it is insufficient to demonstrate the performance advantage of the algorithm on the synthetic datasets. In the future, we will continue exploring and fully exploiting diffusion models' beneficial impact on point cloud pre-training. We also hope that our work will inspire more research on pre-training with diffusion models, contributing to the advancement of the field.

\end{document}